\documentclass{article}
\usepackage{arxiv}
\usepackage[utf8]{inputenc} 
\usepackage[T1]{fontenc}    
\usepackage{hyperref}       
\usepackage{url}            
\usepackage{booktabs}       
\usepackage{amsfonts}       
\usepackage{nicefrac}       
\usepackage{microtype}      
\usepackage{lipsum}		
\usepackage{graphicx}
\usepackage[square,sort,comma,numbers]{natbib}
\usepackage{doi}
\usepackage{amsmath}
\usepackage{subcaption}
\hypersetup{
    colorlinks=true,
    urlcolor=magenta        
}

\title{Class Imbalance in Object Detection: An Experimental Diagnosis and Study of Mitigation Strategies}

\author{ Nieves Crasto \\
	HID \\
	611 Center Ridge Dr\\
	Austin, TX 78753 \\
	\texttt{nieves.crasto@hidglobal.com}
}

\date{}

\renewcommand{\headeright}{Technical Report}
\renewcommand{\undertitle}{Technical Report}
\renewcommand{\shorttitle}{}

\hypersetup{
pdftitle={Class imbalance in object detection: An experimental diagnosis and study of mitigation strategies},
pdfauthor={Nieves Crasto},
pdfkeywords={First keyword, Second keyword, More},
}

\begin{document}
\maketitle

\begin{abstract}
Object detection, a pivotal task in computer vision, is frequently hindered by dataset imbalances,
particularly the underexplored issue of foreground-foreground class imbalance. This lack of attention to foreground-foreground class imbalance becomes even more pronounced in the context of single-stage detectors.
This study introduces a benchmarking framework utilizing the YOLOv5 single-stage detector to address the problem of foreground-foreground class imbalance. We crafted a novel 10-class long-tailed dataset from the COCO dataset, termed COCO-ZIPF, tailored to reflect common real-world detection scenarios with a limited number of object classes. Against this backdrop, we scrutinized three established techniques: sampling, loss weighing, and data augmentation. Our comparative analysis reveals that sampling and loss reweighing methods, while shown to be beneficial in two-stage detector settings, do not translate as effectively in improving YOLOv5's performance on the COCO-ZIPF dataset. On the other hand, data augmentation methods, specifically mosaic and mixup, significantly enhance the model's mean Average Precision (mAP), by introducing more variability and complexity into the training data.\\(Code available: \href{https://github.com/craston/object_detection_cib}{https://github.com/craston/object\_detection\_cib})

\end{abstract}


\section{Introduction}
Object detection, a critical task in computer vision, often grapples with the challenge of imbalance in datasets. This imbalance manifests in various forms such as uneven class distributions \cite{shrivastava2016OHEM} \cite{cao2020prime}, uneven distribution in the size of the bounding boxes \cite{lin2017FPN}, \cite{liu2016ssd}, or even the location of the bounding boxes \cite{rezatofighi2019GIOU}, \cite{zheng2020diou}, \cite{zhang2019freeanchor}, \cite{liu2018PANet} in the images. Further, imbalance in class distributions can be classified into two types: foreground-background \cite{lin2017focalloss} and foreground-foreground. The latter, foreground-foreground class imbalance, is especially critical as it can lead to a skewed detector that favors more commonly occurring objects, potentially at the expense of accurately detecting rarer items.

Extensive research has been dedicated to addressing foreground-background imbalance \cite{li2022equalizedFL}, \cite{cao2020prime}, \cite{lin2017focalloss}, \cite{shrivastava2016OHEM}, \cite{chen2019APloss}, \cite{qian2020dr}, \cite{li2019gradient}, particularly in single-stage detectors where its impact is more pronounced. However, the exploration of foreground-foreground imbalance \cite{ouyang2016factors}, \cite{gupta2019lvis}, \cite{oksuz2020-OFB} in object detection remains relatively underdeveloped, with most existing methods essentially adapting image classification imbalance techniques to the realm of object detection. Furthermore,  these techniques have been largely evaluated in the context of two-stage detectors \cite{gupta2019lvis}, \cite{tang2020badmomentumLVIS}, \cite{wang2020simCAL}, \cite{wang2021seesaw}, \cite{zang2021fasa} leaving a gap in understanding their effectiveness in single-stage models. To bridge this gap we focus on the YOLOv5 (small), a single-stage detector known for its balance of high-speed processing and accuracy, making it ideal for edge deployment for real-time applications. 

Additionally, the performance benchmarks established on academic datasets like COCO \cite{lin2014-COCO} and LVIS\cite{gupta2019lvis}  fail to reflect some of the challenges encountered in practical, real-world scenarios. Models deployed on the edge typically are trained to detect a limited range of object classes, unlike the extensive variety found in standard datasets (80 classes in COCO and over 1200 classes in LVIS). To effectively tackle this issue, we introduce a specialized, smaller dataset, dubbed COCO-ZIPF, which is essentially a long-tailed 10-class curated subset of the COCO dataset. 

 This study presents a thorough benchmarking analysis focused on addressing foreground-foreground class imbalance using the YOLOv5 architecture, evaluating three primary strategies: sampling, loss weighting, and augmentation. Our findings highlight a distinct contrast in effectiveness between these approaches: While sampling and loss weighting have shown promise in improving mean Average Precision (mAP) in two-stage detectors, they prove counterproductive in one-stage detectors like YOLOv5, often diminishing the overall mAP. 
On the other hand, augmentation techniques, particularly mosaic, and mixup, emerge more effective, demonstrating marked improvements in mAP within the YOLOv5 framework. 

\textbf{Our Contributions}:\\ Firstly, we provide a new COCO-ZIPF long-tailed dataset, a 10-class subset of the COCO dataset, to mirror the real-world scenarios of edge-deployed models that detect a constrained set of object classes. Secondly, we establish a PyTorch-based benchmarking framework tailored for YOLOv5 models to assess various strategies combating class imbalance. This framework not only facilitates reproducible model training but also offers a modular setup for integrating alternative datasets and imbalance-mitigation techniques. It also provides support for YOLOv5 - nano models, optimizing it for edge deployment scenarios. Finally, our research underscores the effectiveness of data augmentation methods, particularly mosaic and mixup, which our results show to significantly enhance mean Average Precision (mAP), outperforming sampling and loss weighting when applied to YOLOv5 models trained on our COCO-ZIPF dataset.

\section{Methodology}
\subsection{Dataset Construction}
\begin{figure}[ht]
    \centering
    \begin{subfigure}[b]{0.49\textwidth}
        \includegraphics[width=\textwidth]{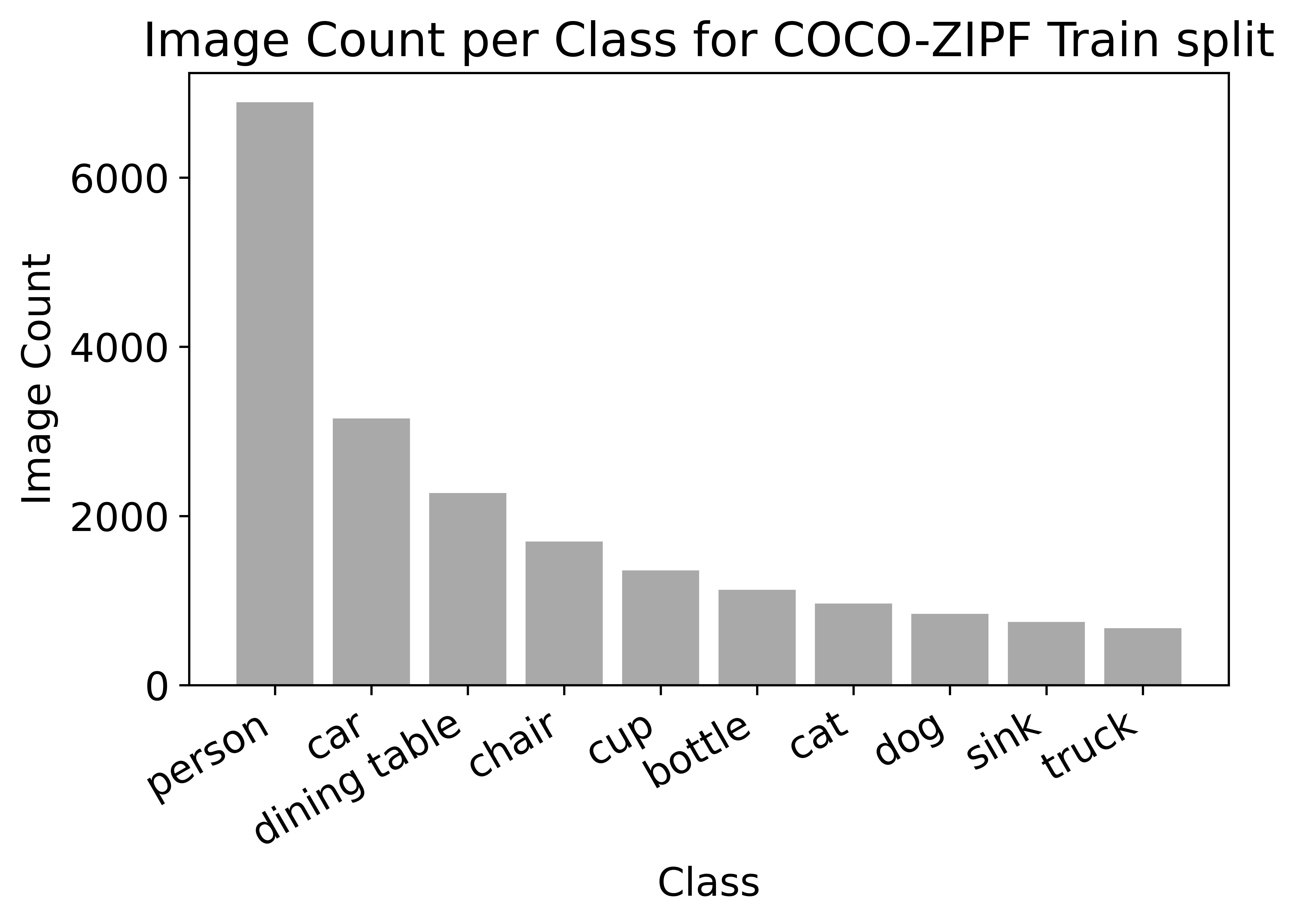}
        \caption{}
        \label{fig:coco_zipf_images_hist}
    \end{subfigure}
    \hfill
    \begin{subfigure}[b]{0.49\textwidth}
        \includegraphics[width=\textwidth]{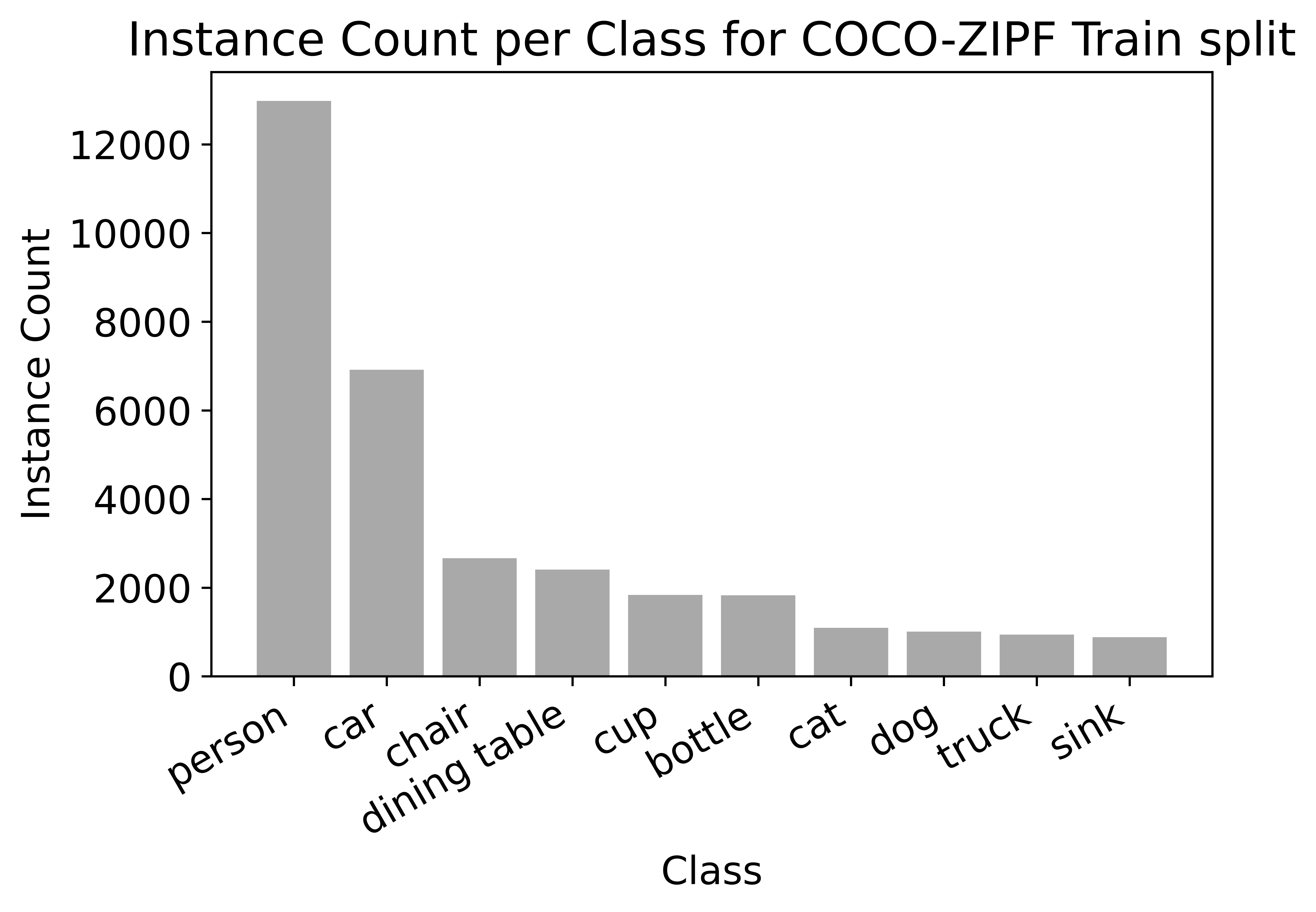}
        \caption{}
        \label{fig:coco_zipf_instances_hist}
    \end{subfigure}
    \caption{Comparative visualization of class instance distributions in the COCO-ZIPF dataset. The left chart (a) displays the count of image instances per class. The right chart (b) represents the number of instances per class. Both charts exhibit a Zipfian distribution, highlighting the long-tail effect in class representation within the dataset.}
    \label{fig:combined}
\end{figure}

To mirror practical detection scenarios, we constructed a condensed dataset named COCO-ZIPF, extracted from the comprehensive COCO-2017 \cite{lin2014-COCO} dataset. This dataset concentrates on the 10 most frequently encountered categories: person, car, dining table, chair, cup, bottle, cat, dog, sink, and truck. We intentionally shaped the dataset to display a long-tailed distribution, with 'person' as the prevalent class and 'truck' as the least represented. We restricted our dataset to images with fewer than 10 object detections for simplicity, ensuring the number of images for each class adhered to a Zipfian distribution (see Equation \ref{eq: zipf}) - a statistical distribution where a few classes have a high number of instances while most have very few, a common pattern in natural datasets and human languages. (see Figure \ref{fig:coco_zipf_images_hist} and Figure \ref{fig:coco_zipf_instances_hist}).

The Zipfian distribution is given by the equation:

\begin{equation}
P(n) = \frac{1/n^s}{\sum_{k=1}^{K} (1/k^s) }
\label{eq: zipf} 
\end{equation}

where \( P(n) \) is the probability of the \( n \)-th most frequent element, \( n \) is the rank, \( s \) is the exponent characterizing the distribution, and \( K \) is the number of elements in the distribution.


\subsection{Model Architecture}
Object detectors are mainly of two types: two-stage and one-stage detectors. Two-stage detectors like Faster RCNN \cite{ren2015faster} and its variants \cite{cai2018cascadeRCNN}, \cite{he2017mask} typically employ a Region Proposal Network (RPN) to generate proposals in the first stage that are then refined and classified in the second stage. On the other hand, one-stage detectors \cite{liu2016ssd}, \cite{redmon2016yolov1}, \cite{redmon2017yolo9000}, \cite{bochkovskiy2020yolov4}, \cite{redmon2017yolo9000}, \cite{lin2017focalloss}

The evolution of the YOLO  series of object detection models has been marked by significant improvements with each iteration. YOLOv1 \cite{redmon2016yolov1} to YOLOv4 \cite{bochkovskiy2020yolov4} utilized a Darknet framework, which was a custom deep learning framework written in C and CUDA. YOLOv5 \cite{yolov5}, however, transitions to a PyTorch framework \cite{paszke2019pytorch}, offering better compatibility with various tools and libraries, and simplifying the deployment pipeline.

The YOLOv5 architecture is split into 3 components: backbone, neck and head. The backbone is based on CSPDarknet53 architecture \cite{bochkovskiy2020yolov4} which uses the CSPNet \cite{wang2020cspnet} strategy to partition the feature map of the base layer into two parts and then merges them through a cross-stage hierarchy. The use of a split and merge strategy allows for more gradient flow through the network. The neck of YOLOv5 incorporates the Spatial Pyramid Pooling Fusion (SPPF) which has been engineered to be computationally more efficient than Spatial Pyramid Pooling introduced in \cite{he2015SPP}. It is reported to be twice as fast as the traditional SPP \cite{yolov5}. The head is based on the YOLOv3 head \cite{redmon2018yolov3}.  
We adopted the YOLOv5 \cite{yolov5} (small) model for our experiments, owing to its compact yet effective design. The model was trained from scratch, incorporating mosaic augmentation to enrich the training dataset. This setup served as our baseline for evaluating the impact of the different class balancing strategies. 

\subsection{Methods dealing with Foreground-Foreground Class Imbalance}
In our research, we focus on evaluating three predominant strategies to address the issue of foreground-foreground class imbalance in object detection: sampling, loss weighting, and augmentation. For the sampling aspect, our analysis encompasses class-aware sampling \cite{shen2016relay}, which strategically balances the representation of various classes in training batches, and repeat factor sampling \cite{gupta2019lvis}, a method designed to give underrepresented classes more representation in the training process. Regarding loss weighting, our approach involves adjusting the weights of the loss function based on the frequency of class instances within the COCO-ZIPF dataset.
Finally, we explore the effectiveness of augmentation techniques, specifically mosaic \cite{bochkovskiy2020yolov4} and mixup\cite{zhang2017mixup}, in improving the model's mean Average Precision (mAP). These augmentation methods are known for their ability to introduce variability and complexity into the training data, which can be beneficial in training models to generalize better.

\subsubsection{Sampling}
Object detection datasets typically contain multiple instances of objects within a single image with over-represented objects often co-occurring with under-represented classes. This characteristic can lead to a situation where oversampling techniques designed to increase the prevalence of underrepresented classes also increase the number of instances of majority classes present in the oversampled images. For example, in urban scenes, 'cars' are typically pervasive and are commonly seen with 'traffic lights' or 'pedestrians'. When applying oversampling to increase the visibility of 'traffic lights', there is an unintended consequence of also multiplying the instances of 'cars'. Despite this, techniques such as class-aware sampling \cite{shen2016relay} and repeat-factor sampling \cite{gupta2019lvis} have yielded modest gains in mean Average Precision (mAP) for two-stage detectors. We are now keen to ascertain whether these enhancements can be effectively translated into the YOLOv5 framework.

\subsubsection*{Class-Aware Sampling}
Class-aware sampling (CAS) is a technique aimed at addressing class imbalance in object detection datasets. This method, initially proposed by \cite{shen2016relay}, focuses on optimizing the composition of mini-batches during the training process to ensure a uniform representation of classes. 

The technique can be summarized as follows:

\begin{itemize}
    \item A list of classes and a corresponding list of images for each class are maintained.
    \item During training, mini-batches are constructed by first randomly selecting a class and then randomly selecting an image containing an instance of that class.
\end{itemize}

This approach ensures that each class, irrespective of its frequency in the dataset, has an equal chance of being represented in the training batches. While CAS can enhance the model's exposure to minority classes, it may inadvertently reduce the frequency of the majority class in the training batches. Therefore, it is crucial to monitor the model's performance on both minority and majority classes to maintain a balance.

\subsubsection*{Repeat Factor Sampling}
Repeat Factor Sampling (RFS) is a strategy designed to address the class imbalance issue in long-tailed object detection datasets. This technique, as described by \cite{gupta2019lvis}, involves oversampling images containing instances of under-represented classes. The core idea is to calculate a repeat factor for each image, dictating the frequency of its presence in the training set, thereby ensuring enhanced representation of rarer classes.

Given a dataset, the repeat factor for each image $i$, denoted as $r_i$, is calculated based on the presence of different object classes within that image. The process is as follows:

For each class $c$, we first determine its frequency $f_c$, defined as the ratio of the number of training images containing class $c$ to the total number of training images:

\begin{equation}
    f_c = \frac{\text{Number of training images containing class } c}{\text{Total number of training images}} \nonumber
\end{equation}

The category-level repeat factor $r_c$ for each class $c$ is then computed using a threshold parameter $t$:

\begin{equation}
    r_c = \max\left(1, \sqrt{\frac{t}{f_c}}\right) \nonumber
\end{equation}

where $t$ is a hyperparameter that determines the threshold for oversampling. For categories with $f_c \leq t$, there is significant oversampling.

The image-level repeat factor for each image $i$, $r_i$, incorporates the repeat factors of all classes present in the image. It is calculated as the maximum (or the mean) of the repeat factors of these classes:

\begin{equation}
    r_i = \max_{c \in i} r_c \quad \text{or} \quad r_i = \text{mean}_{c \in i} r_c \nonumber
\end{equation}

where $\{c \in i\}$ represents the set of classes present in image $i$.

RFS adjusts the sampling probability of each image during training, such that images with higher repeat factors are more likely to be selected. This approach effectively increases the prevalence of under-represented classes in the training process, aiming to achieve a more balanced model performance across all classes. The implementation of RFS requires careful tuning of the threshold parameter $t$, as it directly influences the degree of oversampling for rare classes. In our experiments, we set $t=1.0$.

\subsubsection{Loss Reweighing}
The second strategy involved modifying the loss function to account for class frequencies. By adjusting the weights of losses for each class, the model's learning was directed to equally focus on all classes, preventing the dominance of any single class. Yolov5 uses binary cross entropy as the classification loss (see Equation \ref{eq:BCE}). 
\begin{equation}
    \text{Binary Cross-Entropy Loss} = -\frac{1}{N}\sum_{i=1}^{N} [y_i \cdot \log(p_i) + (1 - y_i) \cdot \log(1 - p_i)] 
    \label{eq:BCE}
\end{equation}
where $N$ is the number of observations, $y_i$, is the true label for each observation, which can be either 0 or 1 and $p_i$ is the predicted probability of the observation being in class 1.

We can also incorporate weights in Equation \ref{eq:BCE} to take into account the class imbalance.
\begin{eqnarray}
\text{Weighted Binary Cross-Entropy Loss} &=& -\frac{1}{N}\sum_{i=1}^{N} \left[ w_{c} \cdot y_i \cdot \log(p_i) +  \cdot (1 - y_i) \cdot \log(1 - p_i) \right] \label{eq:weighted BCE 1}\\  
w_c &=& \dfrac{\text{Total number of instances across all classes}}{ \text{Total number of instances for class } c} \label{eq:weighted BCE 2}
\end{eqnarray}

\subsubsection{Augmentation}
We investigate two prevalent augmentation strategies, mosaic and mixup, widely acknowledged for their regularization benefits in computer vision tasks. These techniques have been integral in enhancing model robustness and performance. Additionally, the copy-paste technique has shown potential for performance improvement, albeit with a significant limitation—the prerequisite of segmentation masks. Given the labor-intensive nature of manual data collection and annotation in real-world scenarios, the generation of segmentation masks becomes a strenuous and time-consuming task. Consequently, to align our study with practical applicability and efficiency, we have excluded the copy-paste technique from our evaluation within this research.
\subsubsection*{Mosaic Augmentation}
Mosaic augmentation, introduced in \cite{bochkovskiy2020yolov4}, is designed to improve the robustness and 
generalization of object detection models. It involves creating a composite image by stitching together four different training images. This process can be described as follows:
Given four training images, $(x_1, x_2, x_3, x_4)$, the Mosaic image $\tilde{x}$ is created by first dividing a canvas into four quadrants. Each quadrant is then filled with a cropped region from one of the training images. The size and position of the cropped region from each image are determined randomly but constrained so that the central region of the composite image contains overlapping parts of all four images. The corresponding labels $\tilde{y}$ are also adjusted to reflect the new positions of the objects.

Mathematically, if $(x_i, y_i)$ represents the $i^{th}$ training image and its label, and $P_i$ denotes the cropping and placement function for the $i^{th}$ image, the Mosaic image $\tilde{x}$ and its label $\tilde{y}$ can be represented as:

\begin{equation}
    \tilde{x} = \begin{bmatrix}
        P_1(x_1) & P_2(x_2) \\
        P_3(x_3) & P_4(x_4)
    \end{bmatrix} \nonumber
\end{equation}
\begin{equation}
    \tilde{y} = \text{AdjustLabels}(P_1(y_1), P_2(y_2), P_3(y_3), P_4(y_4)), \nonumber
\end{equation}

where $\text{AdjustLabels}$ is a function that modifies the labels to reflect the transformed coordinates of objects in the Mosaic image. This augmentation strategy enhances the model's exposure to a variety of object scales, aspect ratios, and contextual information, which is crucial for improving detection performance in complex real-world scenarios.

\subsubsection*{Mixup Augmentation}
Mixup augmentation \cite{zhang2017mixup} is a technique that generates new training samples by linearly combining pairs of existing samples and their labels. The mathematical formulation for mixup is as follows:
Given two training samples, $(x_i, y_i)$ and $(x_j, y_j)$, and a mixing coefficient $\lambda$ drawn from a Beta distribution, $\text{Beta}(\alpha, \alpha)$ for some $\alpha > 0$, the mixup samples $(\tilde{x}, \tilde{y})$ are created as:

\begin{equation}
    \tilde{x} = \lambda x_i + (1 - \lambda) x_j \nonumber
\end{equation}

In this context, $x_i$ and $x_j$ are the input images, while $y_i$ and $y_j$ are their corresponding label vectors. The parameter $\alpha$ controls the strength of interpolation between pairs of samples; a higher value of $\alpha$ leads to stronger mixing. The output labels $\tilde{y}$ is the concatenation of the two input labels $y_i$ and $y_j$.

In our specific application, we employ this technique utilizing pairs of images that have already been enhanced through mosaic augmentation to create our mixup augmented samples. By employing mixup augmentation, the model is exposed to a more diverse set of training examples, which can help in improving its generalization capabilities and mitigating the effects of class imbalance.

\section{Implementation Details}
\subsection{Dataset and metrics}
The COCO-ZIPF dataset was engineered using FiftyOne \cite{moore2020fiftyone}, a versatile open-source dataset management toolkit. Utilizing the train and validation splits from the COCO-2017 dataset available in the FiftyOne dataset zoo, we constructed our own training and validation subsets for COCO-ZIPF. Given the requirement to train a YOLOv5 (small) model from the ground up, we aimed for a training set of approximately 20,000 images. Recognizing that the selection process might filter out numerous images when narrowing down from 80 to 10 classes and establishing a long-tailed distribution, we preemptively sourced a fourfold larger pool from the COCO-2017 training set, totaling 80,000 images, alongside the complete validation set of COCO-2017. Initially, we eliminated images with more than 10 detections from both the training and validation sets to simplify the detection task, particularly to avoid the complexity of detecting numerous small objects within a single frame. For the training subset, we identified and retained the top-10 classes based on the highest image counts. Subsequently, we purged detections belonging to the residual 70 classes from both the train and validation subsets. To model a long-tailed distribution within the training set, we employed a Zipfian distribution with $s=1.01$, as outlined in Equation \ref{eq: zipf}. With the distribution defined, we meticulously filtered surplus images, prioritizing the retention of the under-represented classes. This filtration yielded a training set of roughly 12,000 images, with instance counts as follows: 'person': 12,982, 'car': 6,918, 'chair': 2,663, 'dining table': 2,409, 'cup': 1,837, 'bottle': 1,829, 'cat': 1,096, 'dog': 1,009, 'truck': 941, 'sink': 883 (See Figure \ref{fig:coco_zipf_instances_hist}. For the validation set, we were left with approximately 2,000 images with instance counts detailed as: 'person': 3,307, 'car': 604, 'chair': 380, 'bottle': 314, 'dining table': 240, 'cup': 222, 'truck': 201, 'cat': 188, 'dog': 186, 'sink': 173.

\textbf{Metrics: } For all our results, we report the mean average precision (mAP) over the COCO-ZIPF validation set.

\subsection{Framework Details}
Our framework, designed for advancing object detection research, is built on PyTorch Lightning \cite{Falcon_PyTorch_Lightning_2019}, a high-level interface for PyTorch. PyTorch Lightning enhances the coding process by streamlining complex network training into more manageable and cleaner code. This choice promotes better organization, boosts reproducibility, and accommodates scalability, crucial for experimenting with various object detection models like YOLOv5. For configuration management, we integrate Hydra \cite{Yadan2019Hydra}, a powerful Python framework. Hydra excels in managing complex configurations, enabling us to dynamically compose and override hierarchical configurations via both configuration files and command-line arguments. This flexibility is invaluable in experimenting with different hyperparameter settings, allowing for easy adjustments and fine-tuning of the models without altering the codebase. Furthermore, our framework incorporates advanced logging capabilities. It supports the generation of detailed CSV files for tracking logs and parameters, crucial for analyzing the training process and model performance. Additionally, we offer integration with Weights \& Biases \cite{wandb}, a comprehensive machine learning platform to provide a seamless experience for experiment tracking, and workflow management. 

Overall, the combination of these advanced tools in our framework provides a robust foundation for object detection research. It not only simplifies the development and training of complex models but also ensures that the research process is streamlined, reproducible, and scalable, aligning with the rigorous demands of cutting-edge computer vision research.

\subsection{Training Setup}
We chose the YOLOv5 small (henceforth denoted as YOLOv5s or yv5s) model for evaluation due to its small size, high speed, and decent accuracy which makes it an ideal candidate for edge deployment. The model is trained from scratch on NVIDIA GeForce RTX 3090 GPU with a batch size of $64$ and $8$ workers for $300$ epochs. The images fed into the model were resized to a resolution of $416 \times 416$ pixels. We adopted a linear learning rate scheduling approach, initiating the training with a learning rate of $0.01$ and gradually reducing it to $0.0001$. The initial phase of training included a warm-up period spanning $3$ epochs, during which we set a higher bias learning rate of $0.1$ and a momentum of $0.8$. For the optimization algorithm, Stochastic Gradient Descent (SGD) was utilized, configured with a momentum of $0.937$ and a weight decay of $0.0005$.  The binary cross-entropy loss was applied to both objectness and classification tasks. For localization, we used the CIOU loss \cite{zheng2020diou}. The weighting factors assigned to the objectness, classification, and localization losses were $1.0$, $0.5$, and $0.05$, respectively.

\section{Results and Discussion}
Our experimentation involved the implementation and evaluation of each proposed strategy within the YOLOv5 framework. The performance metrics focused primarily on mean Average Precision (mAP) and its variants across different classes. The baseline model established a foundational benchmark for comparison.
\begin{table}[ht]
\small
\centering
\begin{tabular}{|l|r|r|r|r|r|r|r|}
\hline 
\shortstack{Metric} & \shortstack{Nb. of\\instances} & \shortstack{Yv5s \\ no mosaic} & \shortstack{Yv5s + \\ mosaic \\(baseline)} & \shortstack{Yv5s + \\mosaic + \\CAS} & \shortstack{Yv5s + \\mosaic + \\RF(mean)} & \shortstack{Yv5s +\\ mosaic + \\Loss weights (tuned)} & \shortstack{Yv5s + \\ mosaic + \\ mixup(prob =0.3)} \\
\hline
map & & 25.4 & 32.0 & 31.4 (-0.6) & 31.1 (-0.9) & 31.2 (-0.8) & 34.3 (+2.3) \\
map50 & & 35.8 & 43.8 & 43.2 (-0.6) & 42.3 (-1.5) & 41.9 (-1.9) & 47.1 (+3.3) \\
map50\_person & 12982 & 65.5 & 71.6 & 71.1 (-0.5) & 70.7 (-0.9) & 71.2 (-0.4) & 74.1 (+2.5) \\
map50\_car & 6918 & 37.0 & 43.1 & 43.1 (0.0) & 43.5 (-0.4) & 44.4 (+1.3) & 45.4 (+2.3) \\
map50\_dining table & 2409 & 35.7 & 40.5 & 39.4 (-1.1) & 40.5 (0.0) & 38.3 (-2.2) & 40.0 (-0.5) \\
map50\_chair & 2663 & 19.2 & 21.6 & 21.3 (-0.3) & 19.1 (-2.5) & 18.8 (-2.8) & 21.9 (+0.3) \\
map50\_cup & 1837 & 24.0 & 29.3 & 32.5 (+3.2) & 30.2 (-0.9) & 30.8 (+1.5) & 36.9 (+7.6) \\
map50\_bottle & 1829 & 15.6 & 24.5 & 24.8 (+0.3) & 23.8 (-0.7) & 24.8 (+0.3) & 30.1 (+5.6) \\
map50\_cat & 1096 & 57.1 & 72.8 & 71.5 (-1.3) & 68.0 (-4.8) & 69.0 (-3.8) & 76.3 (+3.5) \\
map50\_dog & 1009 & 45.2 & 58.4 & 57.4 (-1.0) & 56.7 (-1.7) & 51.3 (-7.1) & 64.1 (+5.7) \\
map50\_sink & 883 & 34.4 & 41.7 & 38.6 (-3.1) & 40.9 (-0.8) & 39.6 (-2.1) & 45.4 (+3.7) \\
map50\_truck & 941 & 24.4 & 34.0 & 32.8 (-1.2) & 30.1 (-3.9) & 30.7 (-3.3) & 37.2 (+3.2) \\
\hline
\end{tabular}
\caption{Performance metrics for YOLOv5s models under various augmentation and sampling strategies}
\label{tab:performance_metrics}
\end{table}

\subsection{Sampling}
The class-aware sampling method showed mixed results. While it improved detection in some minority classes, it inadvertently reduced the model's accuracy for the majority class. This trade-off was evident in the overall decrease in mAP in comparison to the baseline, indicating the complexity of achieving a balance between class representations.

\begin{figure}[ht]
    \centering
    \begin{subfigure}[b]{0.3\textwidth}
        \includegraphics[width=\textwidth]{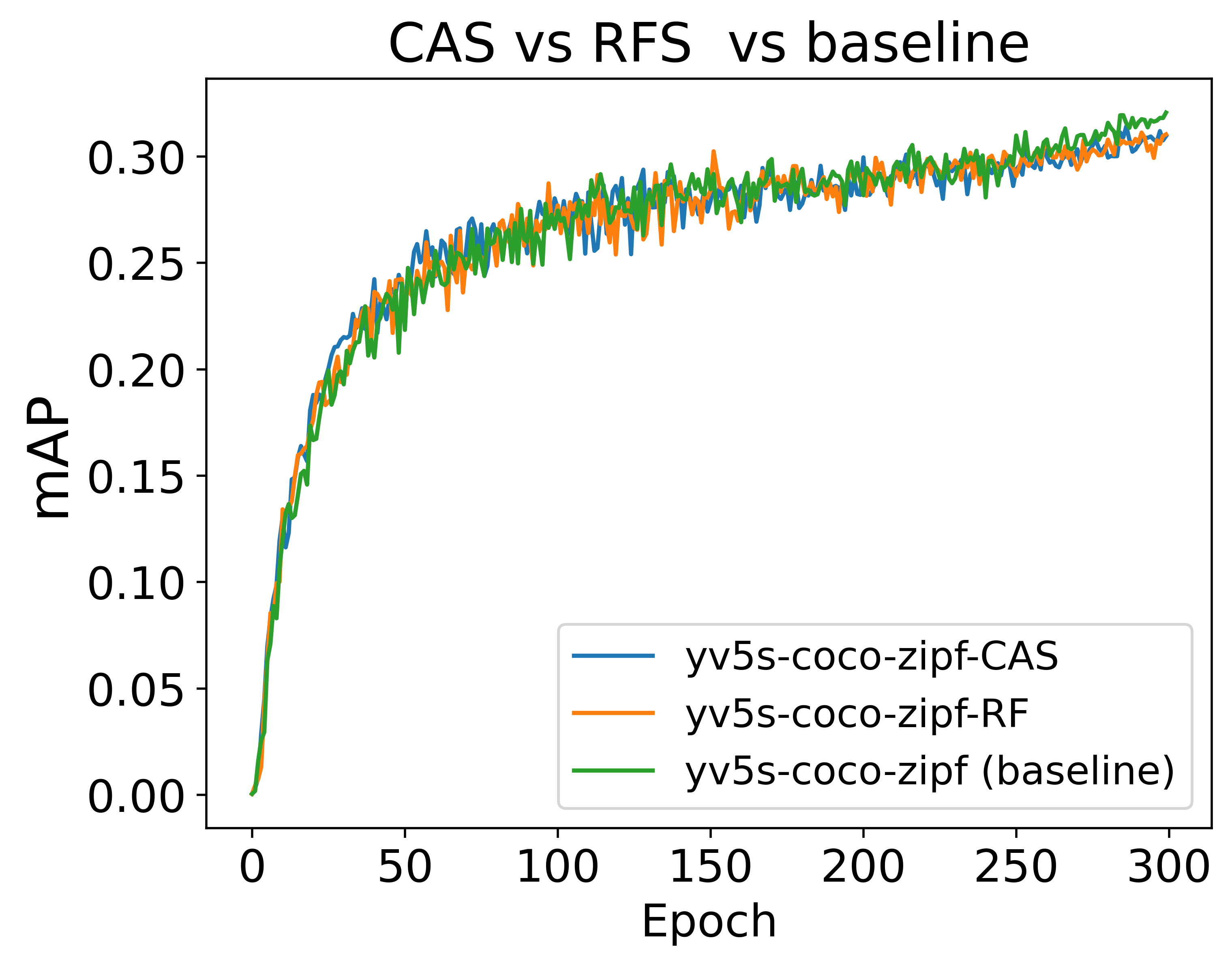}
        \caption{Comparative analysis of mean Average Precision (mAP) over training epochs for YOLOv5s with baseline, Class-Aware Sampling (CAS), and Repeat Factor Sampling (RFS) on COCO-ZIPF }
        \label{fig:coco_zipf_sampling}
    \end{subfigure} \hfill
    \begin{subfigure}[b]{0.3\textwidth}
        \includegraphics[width=\textwidth]{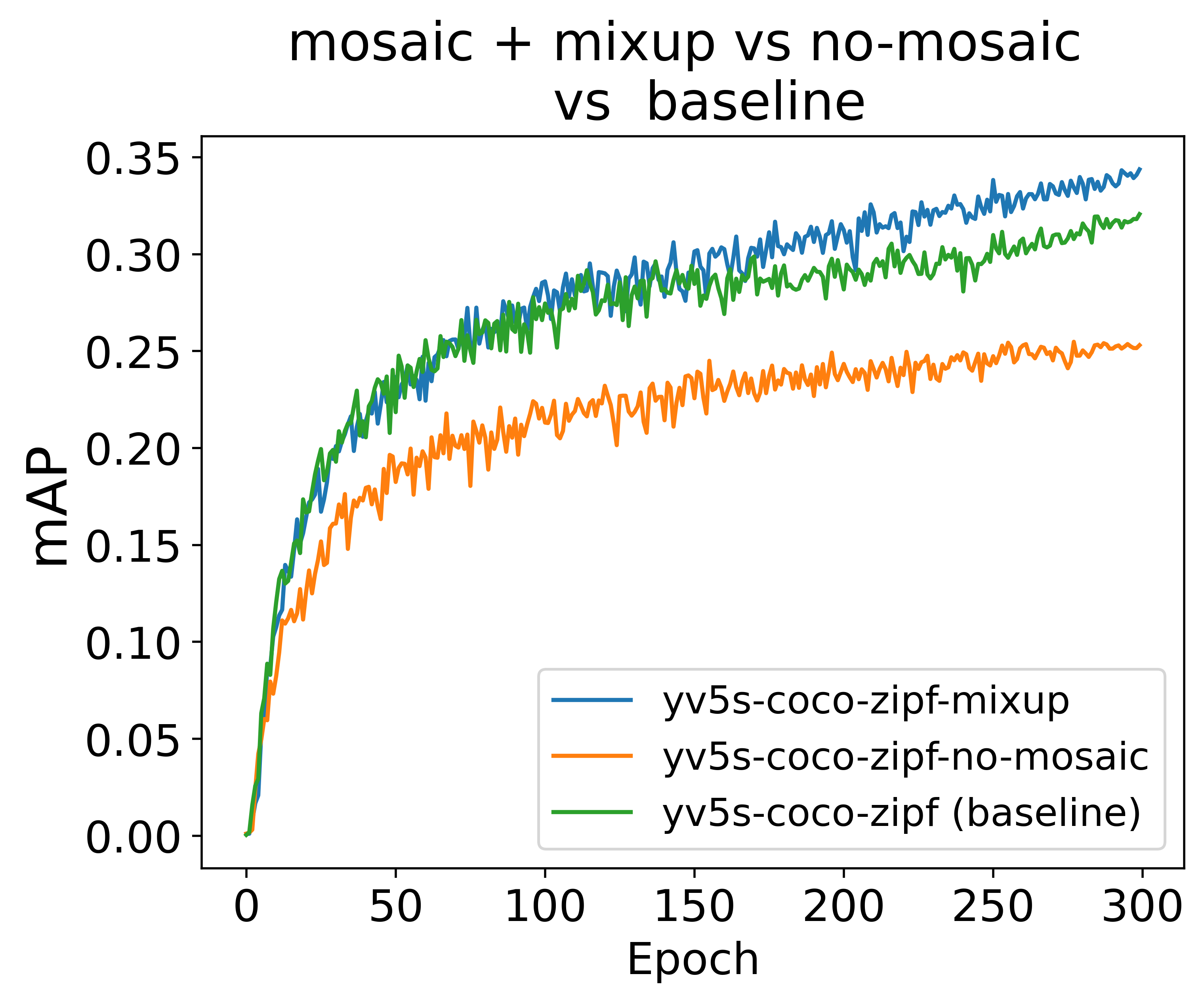}
        \caption{Comparative analysis of mean Average Precision (mAP) over training epochs for YOLOv5s without mosaic, baseline (with mosaic), and with mosaic and mixup (probability of applying mixup = 0.3)}
        \label{fig:coco_zipf_mosaic_mixup_no_mosaic}
    \end{subfigure} \hfill
    \begin{subfigure}[b]{0.3\textwidth}
        \includegraphics[width=\textwidth]{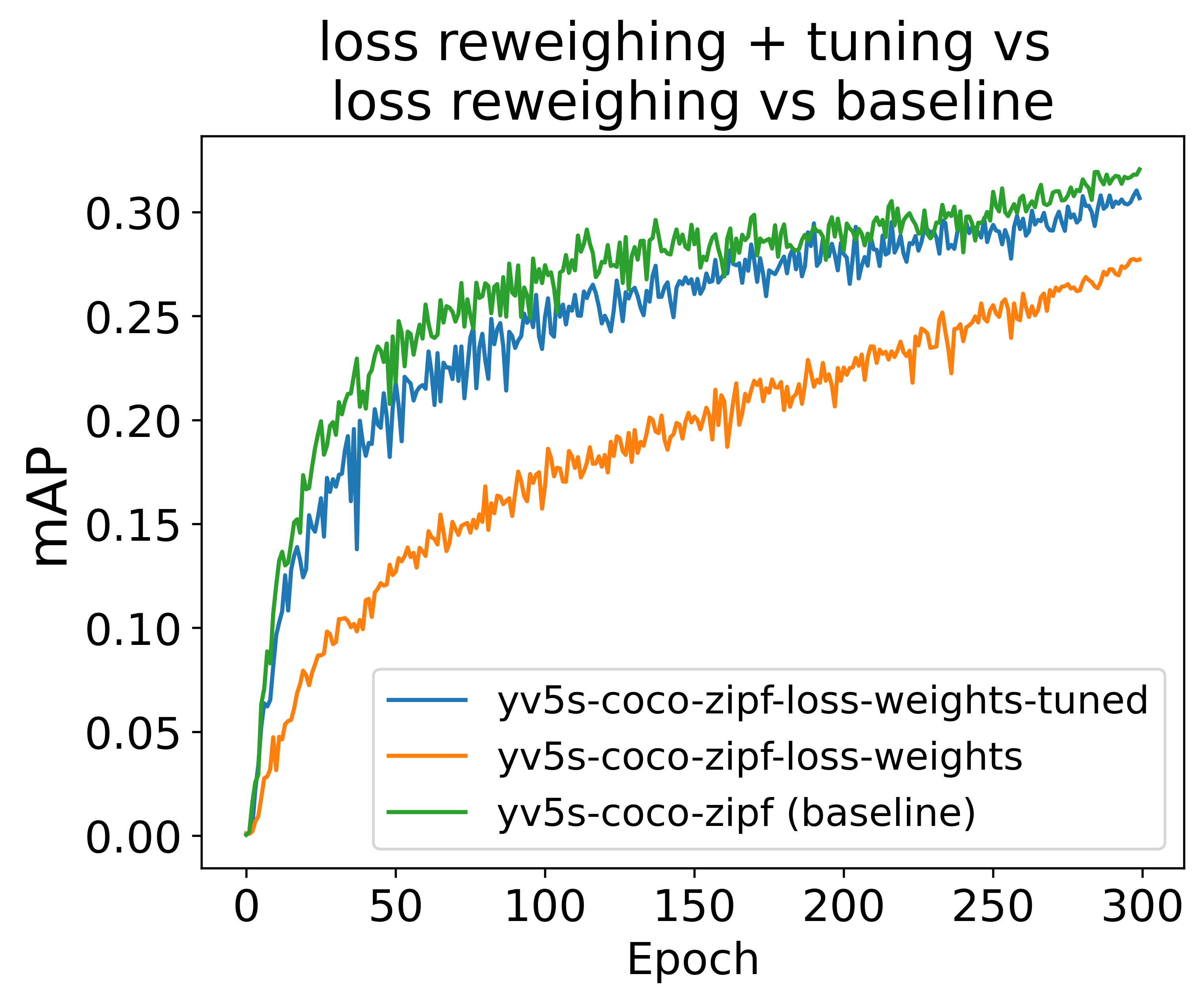}
        \caption{Comparative analysis of mean Average Precision (mAP) over training epochs for YOLOv5s without mosaic, baseline (with mosaic), and with mosaic and mixup (probability of applying mixup = 0.3)}
        \label{fig:loss_weights_map}
    \end{subfigure}
    \\\vspace{3mm}
    \begin{subfigure}[b]{0.33\textwidth}
        \includegraphics[width=\textwidth]{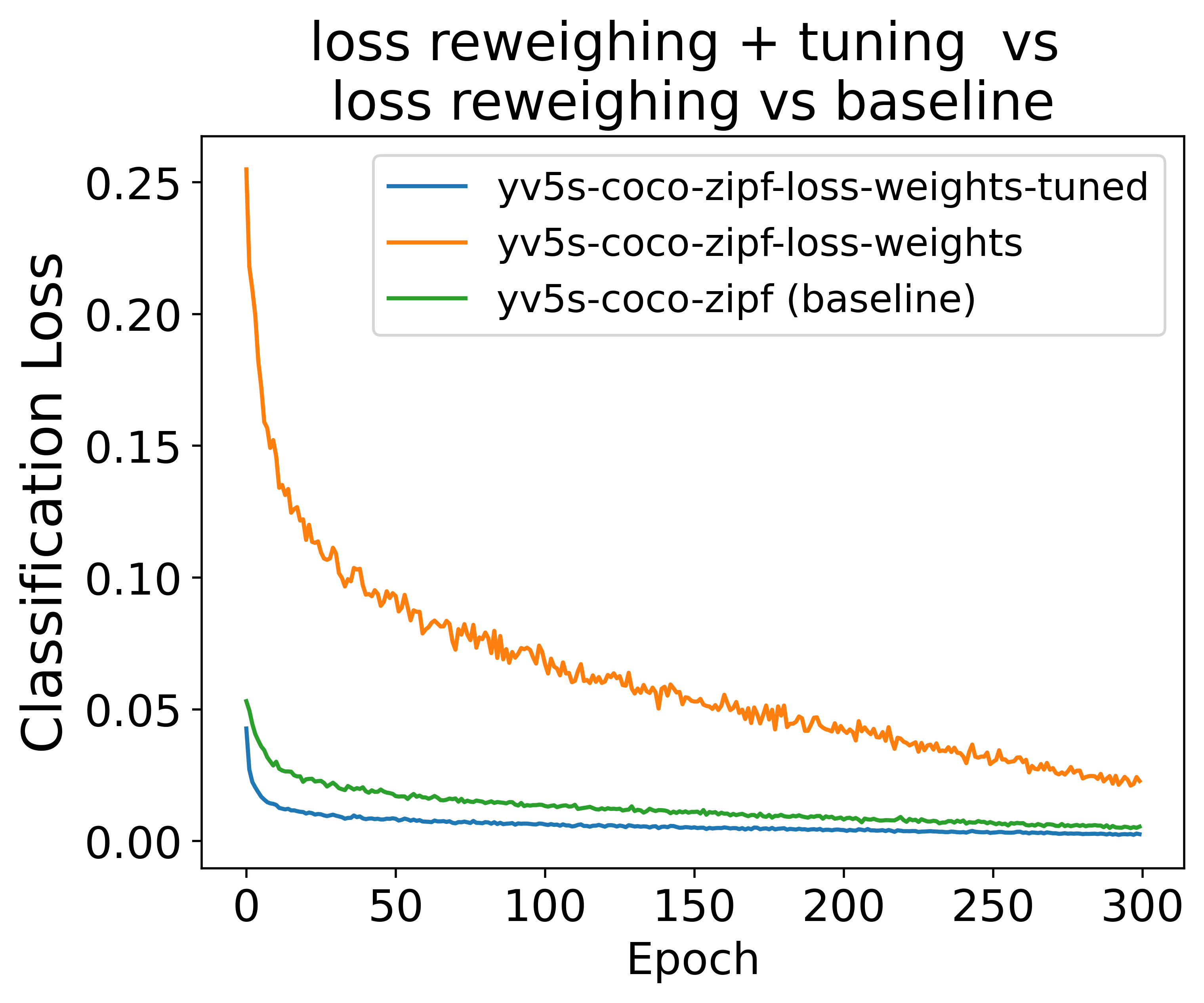}
        \caption{Comparative analysis of Classification Loss over training epochs for YOLOv5s with loss reweighing (with and without tuning) versus the baseline}
        \label{fig:loss_weights}
    \end{subfigure} \hspace{3mm}
    \begin{subfigure}[b]{0.33\textwidth}
        \includegraphics[width=\textwidth]{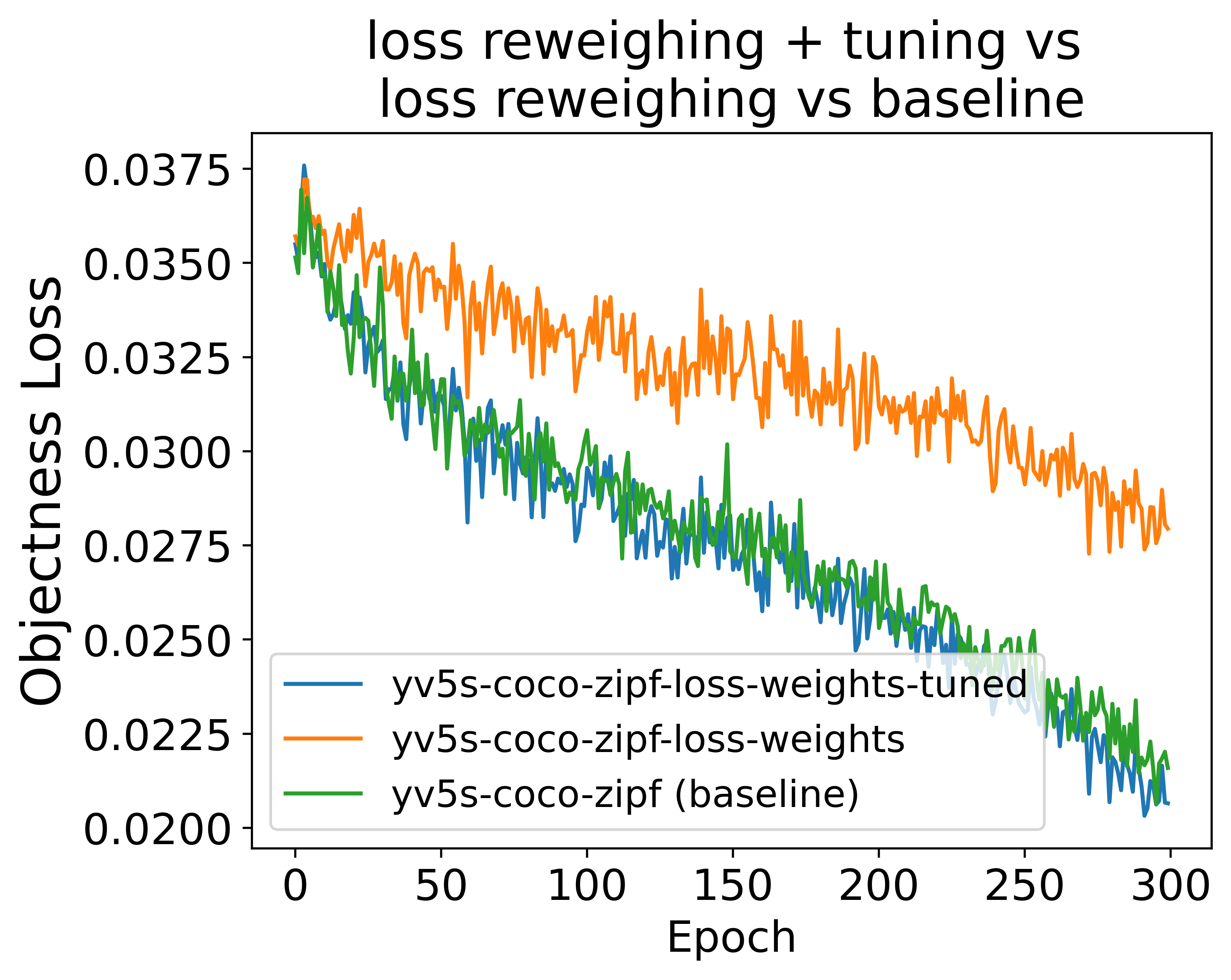}
        \caption{Comparative analysis of Objectness Loss over training epochs for YOLOv5s with loss reweighing (with and without tuning) versus the baseline}
        \label{fig:loss_weights_objectness}
    \end{subfigure}
    
    \caption{}
\end{figure}

The figure [\ref{fig:coco_zipf_sampling}] presents a comparison of mean Average Precision (mAP) over 300 epochs for different sampling strategies applied to the YOLOv5s model on the COCO-ZIPF dataset. The blue line represents the model trained using Class-Aware Sampling (CAS), the orange line indicates Repeat Factor Sampling (RFS), and the green line denotes the baseline YOLOv5s model with usual random sampling.  Throughout the training duration, the mAP trajectories of both the CAS and RFS strategies closely mirror that of the baseline, with a marginal decrease observed towards the final epochs, suggesting that CAS and RFS do not outperform the baseline.

\subsection{Loss Reweighing}
Using the weighted binary cross-entropy loss as defined in equations \ref{eq:weighted BCE 1} and \ref{eq:weighted BCE 2}, we encountered a significant performance degradation as seen the orange line in Figure \ref{fig:loss_weights_map}.
This performance drop indicates a disproportionate influence of different components of the loss function, particularly noticeable in the early stages of training where classification loss predominates over objectness loss (see Figure \ref{fig:loss_weights} and Figure \ref{fig:loss_weights_objectness}). To counteract this imbalance, we recalibrated the weight of the classification loss from 0.5 to 0.05. This adjustment, labeled as yv5s-coco-zipf-loss-weights-tuned (blue line in Figure \ref{fig:loss_weights_map}, helped in improving the mean Average Precision (mAP). 
However, it still falls short of the baseline model's performance (green line in Figure \ref{fig:loss_weights_map}. We opted for a cross-entropy loss function, applying class-specific weights derived from Equation \ref{eq:weighted BCE 2}, to train the model. This method yielded results comparable to those obtained with the weighted binary cross-entropy approach, highlighting the complex nature of optimizing loss functions for improved model performance.

\subsection{Augmentation}
The augmentation strategies, particularly mosaic, demonstrated significant improvements in the model's performance (see Table \ref{tab:performance_metrics}).  Additionally, integrating mixup into mosaic-augmented images with a probability of 0.3 led to a further increase in mAP.  A key observation from our experiments is that both mosaic and mixup augmentations contribute to improved performance uniformly across classes, benefiting both over-represented and under-represented categories. In an attempt to specifically boost the representation of under-represented classes, we modified the mosaic generation process to selectively include images with under-represented classes more frequently. However, this approach resulted in a marginal decrease in the detection accuracy of the over-represented classes. Similarly, with mixup augmentation, we conducted a trial where the second image in the mixup pair was specifically generated from a mosaic comprising primarily of under-represented class images. Conversely, the first image used in the mixup process was created from a mosaic that sampled from the entire dataset.  This method, too, did not yield any significant improvement in overall performance.

Figure \ref{fig:coco_zipf_mosaic_mixup_no_mosaic} demonstrates that the model trained without mosaic augmentation (yv5s-coco-zipf-no-mosaic, depicted by the orange line) persistently records the lowest mean average precision, hinting that the absence of mosaic augmentation could be unfavorable. Conversely, the model employing mixup augmentation (yv5s-coco-zipf-mixup0.3, represented by the blue line) begins its trajectory in line with the baseline model (baseline, indicated by the green line) and eventually surpasses it. This trend suggests that integrating the mixup technique may confer a significant advantage to the model's accuracy.

\section{Conclusion}
In this study, we delved into the challenges of foreground-foreground class imbalance, particularly within the context of YOLOv5 model, single-stage detectors, which is ideally suited for edge computing scenarios. Our approach involved the creation of COCO-ZIPF, a 10-class subset of the COCO dataset, tailored to mirror the typical object detection requirements in real-world edge computing applications, where a limited range of object classes is often sufficient. In order to simplify the YOLO model development and training process, we developed a benchmarking framework to ensure reproducible research. Additionally, our investigation into various techniques to mitigate class imbalance revealed that augmentation strategies, specifically mosaic and mixup, substantially enhance the mAP performance of the YOLOv5 model on our COCO-ZIPF dataset.
\bibliographystyle{plain}
\bibliography{references}

\end{document}